\def\year{2017}\relax
\documentclass[letterpaper]{article}
\usepackage{Template/aaai17}
\usepackage{times}
\usepackage{helvet}
\usepackage{courier}
\usepackage{booktabs}
\usepackage{tabularx}
\usepackage{caption}
\usepackage{listings}
\usepackage{xcolor}
\usepackage{xspace}
\usepackage{paralist}
\usepackage{amsmath,amsfonts,amssymb,stmaryrd}
\usepackage{bigints}
\usepackage{graphicx}
\usepackage{amsthm}
\usepackage{oz}
\usepackage{ifthen}
\usepackage{tabularx}
\usepackage{helvet}
\usepackage{algpseudocode}
\usepackage{algorithm}
\usepackage{algorithmicx}
\usepackage{multirow}
\usepackage{pifont}
\usepackage{natbib}
\usepackage{wasysym}

\algrenewcommand\alglinenumber[1]{\tiny #1:}

\newtheorem{cexpl}{Example}
\newtheorem{cscnr}{Scenario}

\newcommand{\ie}{\mbox{i.\,e.}\@\xspace}

\newcommand{\eg}{\mbox{e.\,g.}\@\xspace}

\newcommand{\secref}[1]{Sec.~\ref{sec:#1}}

\newcommand{\figref}[1]{Fig.~\ref{fig:#1}}
\newcommand{\Figref}[1]{Figure~\ref{fig:#1}}
\newcommand{\tblref}[1]{Tab.~\ref{tbl:#1}}

\newcommand{\explref}[1]{Ex.~\ref{expl:#1}}
\newcommand{\Explref}[1]{Example~\ref{expl:#1}}

\newcommand{\Scnrref}[1]{Scenario~\ref{scnr:#1}}

\lstnewenvironment{limamodel}[1][]
    {\lstset{float=tbp,basicstyle=\footnotesize\ttfamily,#1}}
    {}

\newcommand{\ent}[0]{\ensuremath{\mathcal E} }

\newcommand\multiset[1]{\ensuremath{\llbracket\, #1 \,\rrbracket}}

\newcommand{\densitylabel}[1]{\ensuremath{\text{“}#1\text{”}}}
\newcommand{\density}[2]{\ensuremath{\densitylabel{#1} \mapsto #2}}

\newcommand{\slotn}[1]{\mbox{\fontfamily{phv}\fontseries{mc}\selectfont #1}}
\newcommand{\slot}[2]{\ensuremath{\slotn{#1:\ }\densitylabel{#2}}}

\newcommand{\entity}[2][-1]{
  \ifthenelse{\equal{#1}{-1}}{
    \ensuremath{\langle #2 \rangle}
  }{
    \ensuremath{#1 \langle #2 \rangle}
  }
}

\newcommand{\stateform}[1]{\multiset{#1}}

\newcommand{\context}[1]{\ensuremath{\{\, #1 \,\}}}

\begin{document}

\title{Sequential Lifted Bayesian Filtering in Multiset Rewriting Systems}
\author{Max Schr\"oder, Stefan L\"udtke, Sebastian Bader, Frank Kr\"uger \and Thomas Kirste\\
Mobile Multimedia Information Systems Group, Institute of Computer Science\\
University of Rostock, 18051 Rostock, Germany\\
\{max.schroeder, stefan.luedtke2, sebastian.bader, frank.krueger2, thomas.kirste\}@uni-rostock.de\\
}
\maketitle

\begin{abstract}
Bayesian Filtering for plan and activity recognition is challenging for scenarios that contain many observation equivalent entities (\ie entities that produce the same observations).
This is due to the combinatorial explosion in the number of hypotheses that need to be tracked.
However, this class of problems exhibits a certain symmetry that can be exploited for state space representation and inference.
We analyze current state of the art methods and find that none of them completely fits the requirements arising in this problem class. 
We sketch a novel inference algorithm that provides a solution by incorporating concepts from Lifted Inference algorithms, Probabilistic Multiset Rewriting Systems, and Computational State Space Models.
Two experiments confirm that this novel algorithm has the potential to perform efficient probabilistic inference on this problem class.
\end{abstract}

\section{Introduction}\label{sec:Introduction}

Bayesian Filtering algorithms are often applied in the domain of plan and activity recognition~\citep{sukthankar2008PLpruning,huang09permutation,ramirez2011pomdp}.
These algorithms typically suffer from an exponential state space growth when reasoning about multiple entities that cannot be distinguished from observations.
This is for instance the case in multiple person tracking by use of anonymous sensors~\citep{fox2003bf} or assisted manufacturing, where the identity of different tools and parts is uncertain.
Currently, no algorithm exists that exploits these symmetrical parts of the state space and that is applicable to all instances of the underlying problem class.

Consider the following multiple person tracking example: 
\begin{cexpl}
	Two persons ``A" and ``B" are moving within an environment observed by anonymous presence sensors.
  Both are at the same room.
  When the presence sensor indicates that one person left this room, the following two hypotheses have to be tracked:
\begin{inparaenum}[(1)]
  \item person ``A" left the room and person ``B" stayed in the room, and
  \item the other way round -- ``B" left, ``A" stayed.
\end{inparaenum}
\end{cexpl}

This example is one instance of the \emph{data association problem} sometimes referred to as track confusion problem \citep{fox2003bf, wilson2005tracking, huang09permutation}.
The data association problem occurs if two (or more) entities (\eg persons) are recognized by the same anonymous sensor and thus cannot be distinguished anymore.
In other words, the person's identity associated with the movement track is lost. 
As a consequence, all possible resulting hypotheses have to be tracked.
This leads to a combinatorial explosion with respect to the number of hypotheses that have to be considered.

Tracking single persons or very small groups is feasible using state of the art techniques.
However, tracking more than 10 persons is not computationally feasible without abstraction.
With respect to modeling chemical reaction, Multiset Rewriting Systems (MRS) allow to reason over thousands of atoms by abstracting from individual atoms.
However, applications of activity and plan recognition require the identification of selected individuals.
Thus, MRS are not immediately applicable.

The contribution of this paper is threefold:
\begin{inparaenum}[(1)]
  \item By providing two example scenarios, we analyze the underlying characteristics and specify a general problem class.
  \item We analyze state of the art methods using six evaluation criteria derived from the class.
  \item We sketch a novel inference algorithm that combines ideas from different state of the art methods and present first evaluation results based on our prototypical implementation.
\end{inparaenum}

Next, we further analyze the data association problem and derive evaluation criteria for state of the art methods (\secref{Application_Requirements}).
In \secref{Inference_Methods} we analyze state of the art methods with respect to the evaluation criteria.
As none of the methods is capable of computing all instances of the problem class, we sketch a novel combination of these methods in \secref{LiMa}.
An evaluation of our mechanism using two scenarios is described in \secref{Experiments} and our conclusion and future work in \secref{Conclusion}.

\section{Problem Specification}\label{sec:Application_Requirements}

The data association problem can be found in many domains, \eg monitoring and assistance at manufacturing \citep{aehnelt_information_2015}, assistance of elderly \citep{Muller.Hein:2015}, or monitoring of buildings to support disaster management \citep{krueger2014colleagues}.
An intuition for the characteristics of the underlying problem class is given by the following two scenarios.

\begin{figure}[tb]
  \centering
  \includegraphics[width=0.35\textwidth]{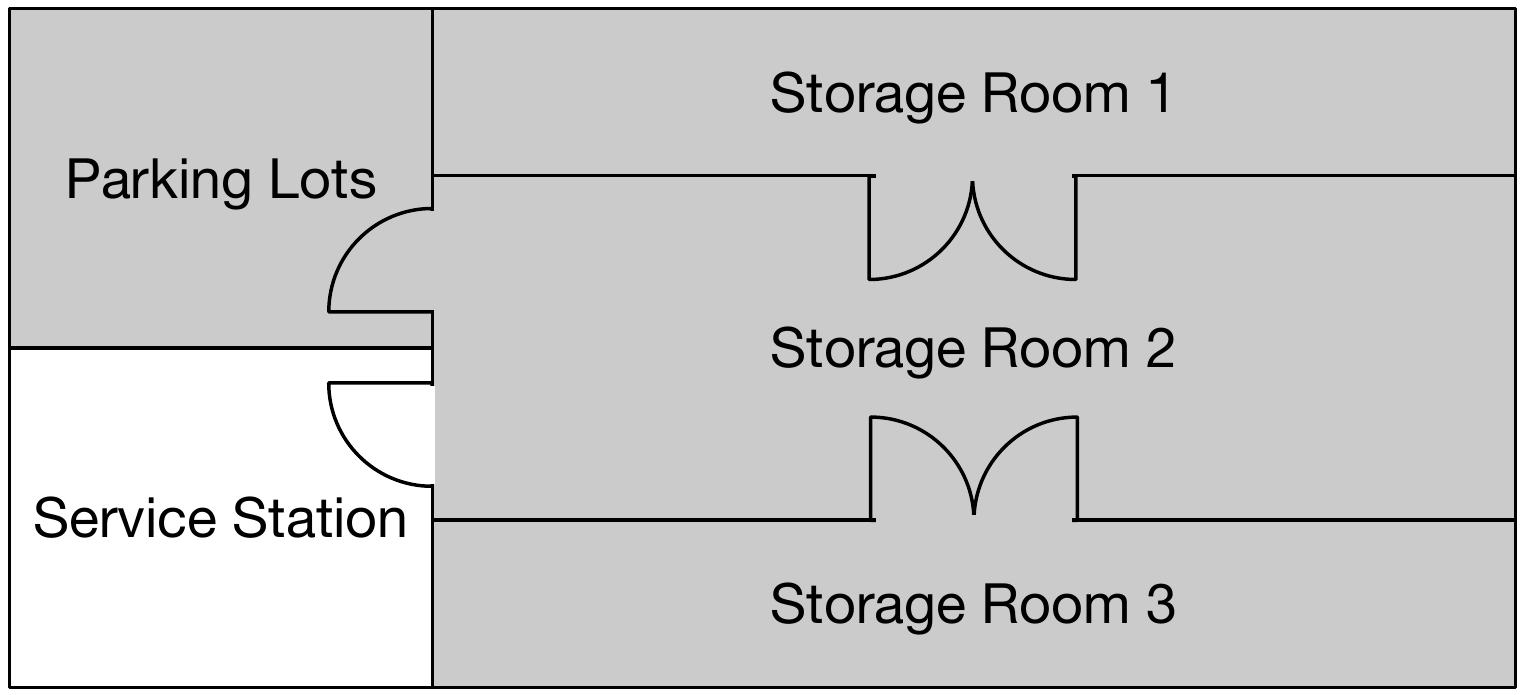}
  \caption{
    Overview of the warehouse scenario.
    Gray areas indicate where multiple forklifts are observation equivalent.
    White areas are equipped with identifying sensors.
  }
  \label{fig:whscenario}
\end{figure}

\begin{cscnr}[Warehouse]\label{scnr:warehouse}
  Ten forklifts are used in a warehouse, consisting of three main storage rooms, a service station where the forklifts get refueled, and a room where all forklifts are parked at night.
  The room layout is depicted in \figref{whscenario}.
  Forklifts can move between two connected locations at each time step.
  Each room is equipped with an anonymous sensor that delivers binary information about whether at least one forklift is present at the room.
  When a forklift gets refueled, it is identified.
  Our goal is to track the position of the forklifts over time to optimize storage management.
\end{cscnr}

Another scenario that does not only track the position of several entities, but also maintains additional information for each entity has been adapted from~\citet{bui2003planrec}:

\begin{cscnr}[Office]\label{scnr:office}
  Three persons act in an office environment which is observed by anonymous sensors.
  The office contains a coffee machine, a coffee jar and a water tap as well as a printer and a paper stack.
  The goal of the agents is to print a number of documents and to get a coffee.
  To do so, the coffee machine needs to be filled with water and ground coffee, and paper must be provided for the printer.
  Whenever a person is operating the printer, she has to authenticate herself and thus can be identified.
  Our goal is to estimate the items a specific person is carrying over time in order to provide assisting operations such as opening doors.
\end{cscnr}

\subsection{General Problem Class}\label{sec:general_problem_class}
The previously described scenarios entail the characteristics of the following general problem class:

\newcommand\cref[1]{C\ref{c:#1}}
\newcommand\clabel[1]{\label{c:#1}}

\begin{compactenum}[{C}1.]
  \item \clabel{entities}
    Multiple entities (possibly of the same kind) are interacting simultaneously to reach their individual goals.
    Entities, here, include passive objects as well as active ones like persons or robots.
    An interaction can be any type of creation, manipulation or deletion of some entities.
    Entities have an identity that might be of interest to answer the application specific question.
  \item \clabel{sensors}
    The scenario is partially observed given a set of (noisy) sensors.
    In the case of anonymous sensors, these do not allow to distinguish between multiple entities as they potentially produce the same sequence of sensor data, \ie they are \emph{observation equivalent}.
    However, there may also occasionally appear sensor measurements that reveal the identity of some entity.
  \item \clabel{prediction}
    Given a situation, future trajectories, \ie sequences of entity interactions towards a future situation, need to be predicted in order to recognize the pursued goal or to provide assistance.
\end{compactenum}

In the warehouse scenario, the entities are forklifts which move simultaneously through the environment and that act autonomously (\cref{entities}).
Their movement is partially observed by binary sensors indicating that one or more forklifts are present at a specific location.
These sensors do not allow to distinguish individual forklifts.
However, identification is provided at the refuel station (\cref{sensors}).
This scenario does not include the need for predicting future action courses.

Considering \Scnrref{office}, the entities are persons moving in the environment as well as devices and consumables.
The entities possibly interact with each other (\eg a person holding paper replenishes the printer) or act independently (\eg by moving to another place).
This interaction is partially observed by anonymous sensors at all locations and identifying sensors at the printer.
In contrast to anonymous sensors, identifying sensors enable the identification of individual entities.
This allows to prune impossible hypotheses (\cref{sensors}).
Predicting future courses of the persons' actions would allow for a situation aware assistance, \eg by opening doors if agents are carrying coffee (\cref{prediction}) and printed documents.

The general problem class includes also other instances that are encoding only some characteristics such as the predator-prey population scenario \citep{gordon2014probprog} or the reaction trajectory of chemical elements \citep{barbuti_roberto_maximally_2011}.

\subsection{Evaluation Criteria}\label{sec:evaluation_criteria}
In \secref{Inference_Methods} we review state of the art approaches which are applicable to these problems.
To evaluate the capabilities of these algorithms, we use the following evaluation criteria arising from \cref{entities}--\cref{prediction}:

\newcommand\eref[1]{E\ref{e:#1}}
\newcommand\elabel[1]{\label{e:#1}}

\begin{compactenum}[E1.]
  \item \elabel{sequential}
  Can an observation sequence be processed sequentially?
  \emph{An observation sequence can be indefinitely long and not be known completely in advance, therefore the algorithm should not depend on the complete sequence.}

  \item \elabel{state_space_change}
    Can the structure of the state space change over time?
    \emph{As interactions of entities may change the set of entities that exist (\cref{entities}), the resulting state space must change accordingly.} 

  \item \elabel{update}
    Is the approach capable of handling nondeterminism resulting from actions and observations?
    \emph{Given a prior distribution, the sensor measurements need to be taken into account for updating this belief state (\cref{sensors}).}

  \item \elabel{prediction}
    Can the approach predict future states, based on the current belief state?
    \emph{This is necessary \eg for goal recognition or to provide assistance.}

  \item \elabel{transitions}
    Are state transitions able to represent multiple parallel actions of individual entities?
    \emph{Entities are interacting simultaneously (\cref{entities}), \ie multiple entities may perform different actions at the same state.}

  \item \elabel{observation_equivalence}
    Can the approach represent equivalent (\eg observation equivalent) aspects of a state compactly when possible and ground this representation when necessary?
    \emph{If parts of the state space cannot be distinguished (\cref{sensors}), the algorithm should combine these to increase efficiency. If an identifying observation is made (\cref{sensors}) or if we are interested in the identity of a specific entity (\cref{entities}), this compact representation needs to be grounded to some extent.}

\end{compactenum}

\section{Related Work and Preliminaries}
\label{sec:Inference_Methods}
Different research domains target different aspects of the problem class. 
Here, we give an overview of techniques related to the problem domain and analyze their strengths and weaknesses with respect to this class.

\subsection{Identity Management (IM)}
Identity Management \citep{kondor_multi-object_2007,huang09permutation} is concerned with multiple-object tracking in case of anonymous observations such as radar measurements.
Objective is to associate observed tracks with concrete entities, that is maintaining a distribution over permutations of entities. 
Identity management is concerned with finding a compact representation of these distributions by approximation employing the first Fourier coefficients of the group of permutations.

This approach allows to perform Bayesian Filtering (BF) (\eref{sequential}, \eref{update}, and \eref{prediction}) to solve the data association problem (\eref{observation_equivalence}). 
It is very efficient and able to handle a large number of entities, \eg up to 100 in \cite{huang_exploiting_2009}.

This ability comes at the cost of a limited expressiveness, as only problems of observing tracks of concrete entities can be modeled.
The approach can not be used in scenarios where additional information about these objects have to be maintained as for instance in the office scenario (\secref{Application_Requirements}).
IM offers no mechanism to model dynamically changing states (\eref{state_space_change}) and parallel actions (\eref{transitions}).

\subsection{Computational State Space Models (CSSM)} 
CSSMs allow the knowledge-based construction of state spaces for BF. 
They are for instance used for human behavior and goal recognition \citep{baker_action_2009,ramirez2011pomdp}.
The transition model is described by a computable function by means of preconditions and effects.
This allows the compact representation of potentially infinite state spaces by avoiding explicit state enumeration, as in Hidden Markov Models.  

CSSMs allow to handle large, even infinite, state spaces \citep{kruger_computational_2014}.
By employing standard methods for BF (\eg Particle Filters), online inference with identifying observations can be performed (\eref{sequential}, \eref{update}, and \eref{prediction}).
A dynamically changing state space (\eref{state_space_change}) can be modeled by state predicates with infinite domain.
CSSMs allow the parallel application of actions of multiple entities (\eref{transitions}).

CSSMs perform inference in grounded state spaces (\ie concrete values are assigned to all state variables).
With respect to observation equivalent states (\eref{observation_equivalence}), CSSM's do not allow any abstraction but rather require to track all hypotheses.
This leads to a combinatorial explosion of the number of states that have to be tracked.

\subsection{Lifted Inference (LI)}
Lifted Inference is concerned with efficient inference in relational graphical models (RGMs). 
An RGM is a compact representation of a graphical model that has certain symmetries: random variables (RVs) that have similar (conditional) distributions are represented as a single \emph{parametric} random variable (par-RV). 
Different exact and approximate methods exist \citep{braz_lifted_2005,kersting_counting_2009,gogate_probabilistic_2016}.

The framework of LI allows the abstract representation of observation equivalent entities via par-RVs (\eref{observation_equivalence}). 
\emph{Split} operations allow to incorporate evidence about particular RVs, \ie dividing a par-RV into the RV that we have evidence about and the remaining RVs.
It is possible to perform probabilistic inference in models with very large domain sizes \citep{poole_first-order_2003,van_den_broeck_lifted_2011}.
 
LI is typically applied to Bayesian Networks, which require the complete observation sequence to be present at the start time. 
Thus, they do not allow for \emph{online inference} --- the recursive computation of the belief state for each time step, without knowing the complete observation sequence in advance (\eref{sequential}, \eref{update}, \eref{prediction}, and \eref{transitions}).
Furthermore, inference is not guaranteed to be linear in the number of time steps.
The algorithms require a fixed network structure, therefore they can not model a dynamically changing state space (\eref{state_space_change}).

\subsection{Lifted Bayesian Filtering (LBF)}
Logical particle filters represent multiple states within one particle. 
This is done by leaving some state predicates undefined \citep{zettlemoyer_logical_2008} or by defining a distribution for some of the predicates, instead of a concrete value.
This implicitly defines a distribution over the states represented by the particle \citep{nitti_particle_2013}.
It is related to the Rao-Blackwellized particle filter \citep{doucet_rao-blackwellised_2000}, as parts of the state are represented by samples and others by sufficient statistics.
Multiple approaches combine LI and BF. 
For example, the interface algorithm \citep{murphy_dynamic_2002} has been extended by using a LI algorithm at each time step \citep{kersting_counting_2009,vlasselaer_exploiting_2016}.

By using the framework of BF, Logical Particle Filters satisfy \eref{sequential}, \eref{update} and \eref{prediction}.
Similar states are grouped by providing distributions over uncertain predicates to implement state space abstraction (\eref{observation_equivalence}).

Although similar states are grouped by providing distributions over uncertain predicates, logical particle filters do not dynamically group similar variables (\eref{observation_equivalence}): Multiple (observation equivalent) entities are always represented by different variables for each entity, that potentially have identical distributions.
Furthermore, none of the models can handle a dynamically changing state structure (\eref{state_space_change}) or has been designed to support parallel actions for state transitions (\eref{transitions}).

\subsection{Multiset Rewriting Systems (MRS)}
Multiset Rewriting Systems are an established formalism for modeling systems with many equal objects.
They are used to model chemical reactions \citep{berry_chemical_1990} or cell interactions \citep{bistarelli_representing_2003}. 
The state is described as a multiset of entities, where each entity is an instance of one of finitely many \emph{species}. 
The reactions of entities are modeled as \emph{multiset rewriting rules} (\eref{prediction}) that have preconditions (entities are consumed by the reaction) and effects (entities are created by the reaction). 
Typically, rewriting rules are executed in parallel (\eref{transitions}), that is, the maximal set of parallel applicable rewriting rules (\emph{maximal parallel step}) is determined. 
There may be more than one maximal parallel step, leading to multiple possible execution paths.

Probabilistic MRS \citep{barbuti_roberto_maximally_2011} are concerned with determining the probability of different execution paths. 
This is done by assigning \emph{rates} to each rewriting rule, which leads to a probability of actions and thus, for each maximal parallel step. 
A set of rewriting rules and an initial multiset defines a probabilistic state space.

Probabilistic MRS allow an abstract representation of a state space via the multiset representation (\eref{observation_equivalence}).
Furthermore, MRSs provide a semantics for parallel action execution (\eref{transitions}) and a process for drawing samples (\eref{prediction}) of the sequential process (\eref{sequential}).
The state space is latently infinite, as an arbitrary number of entities can exist (\eref{state_space_change}).

BF algorithms for MRSs that incorporate observations (\eref{update}) are not available.
MRSs do not maintain identities for entities (\eref{observation_equivalence}).
Furthermore, despite the latently infinite state space, the structure of the state space cannot be extended arbitrarily, as the existing species are defined beforehand (\eref{state_space_change}).

\subsection{Summary}
As evident from the discussion above, none of the methods satisfies all of the requirements E1--E6 (see \tblref{requirements}).
To address the problems described above, a combination of CSSMs, LI and Probabilistic MRSs seems reasonable.

\newcommand*{\greysquare}{\textcolor{gray}{\blacksquare}}
\begin{table}
  \begin{center}
    \begin{tabularx}{\columnwidth}{Xlllllll}
      \toprule
      Method &\eref{sequential}&\eref{state_space_change} & \eref{update}& \eref{prediction} &  \eref{transitions} & \eref{observation_equivalence} &   \\
      \midrule
      IM    & $\CIRCLE$ &$\Circle$ & $\CIRCLE$ & $\CIRCLE$ & $\Circle$  & $\CIRCLE$  \\
      CSSM & $\CIRCLE$ &$\CIRCLE$ & $\CIRCLE$ & $\CIRCLE$ & $\CIRCLE$  & $\Circle$  \\
      LI    & $\Circle$ &$\Circle$ & $\CIRCLE$ & $\Circle$ & $\Circle$  & $\CIRCLE$  \\
      LBF   & $\CIRCLE$ &$\Circle$ & $\CIRCLE$ & $\CIRCLE$ & $\Circle$  & $\LEFTcircle$  \\
      MRS  & $\CIRCLE$ &$\LEFTcircle$ & $\Circle$ & $\CIRCLE$ & $\CIRCLE$  & $\LEFTcircle$  \\
      \bottomrule
    \end{tabularx}
  \end{center}
  \caption{Evaluation of inference methods based on criteria from \secref{Application_Requirements}. $\CIRCLE$: satisfies criterium; $\Circle$: does not satisfy criterium; $\LEFTcircle$: satisfies criterium partially}
  \label{tbl:requirements}
\end{table}

\section{Lifted Marginal Filtering}\label{sec:LiMa}

Our approach -- called \emph{Lifted Marginal Filtering} (LiMa) -- combines different ideas from the approaches described above.
The general concept is inspired by Lifted Inference algorithms, that is, finding a parametric representation for equivalent aspects of the state instead of explicitly enumerating all possibilities.
It is based on defining an update operation to Probabilistic MRSs, or, from a different point of view, introduces a multiset-based state description to CSSMs.
Furthermore, the approach defines \emph{structured} entities, as opposed to atomic entities in MRSs, to be able to incorporate additional information about entities such as the identity.

\subsection{Abstract State Representation}

Like in MRS, similar entities are grouped to overcome the combinatorial explosion.
However, in order to group entities that are similar but not completely equal, the plain multiset representation known from MRS has to be extended to cover the uncertainty in the entity properties.
To this end, the representation of entities is separated into the internal structure of the entity and the actual values.
Entities with the equal structure are maintained in a multiset (in the following called \emph{state formula}).
The \emph{context} describes distributions over values within these structures.
Thus, we get a parametrized description of property value maps.
This concept is similar to that of lifting RV into par-RV in LI.
Next, we will refine the concepts of state formulae and contexts.

A state formula is a multiset of entities. 
An entity, is a map of slots to distribution labels.
Thus, labels are used to map properties to possible property values.
Note that slots of an entity are assumed to be independent from each other.

\begin{cexpl}\label{expl:entity}
  An entity modeling the forklift with two properties location \slotn{loc} and unique identifier \slotn{ID} pointing onto value labels \densitylabel{LStor1} respectively \densitylabel{LID} is denoted as:
  \begin{align*}
    &\entity{\slot{loc}{LStor1}, \slot{ID}{LID}}
  \end{align*}
\end{cexpl}

\begin{cexpl}\label{expl:state_formula}
  A state formula with 9 entities as in \explref{entity} and another entity in storage room 2 can be denoted as:
  \begin{align*}
    \stateform{
      &\entity[9]{\slot{loc}{LStor1}, \slot{ID}{LID}},
      \entity[1]{\slot{loc}{LStor2}, \slot{ID}{LID}}
    }
  \end{align*}
\end{cexpl}

The state formula is representing the structure of the entities only.
It is connected to the context via distribution labels.
A context is a map from distribution labels to distribution representations.
In order to employ the same distribution in different entities, the same distribution label has to be referred to.
Note that this allows us to represent slot values of multiple entities that depend on each other.
This is for instance done in \explref{state_formula}, where the identifier ($\slotn{ID}$) of all entities is distributed according to $\densitylabel{LID}$ (thus making sure that no two entities have the same name).

\begin{cexpl}
  Let $\mathcal{U}(fl1,fl2,\dots,fl10)$ be the representation of the distribution that represents an urn without replacement and contains ten identifiers: $fl1$, $fl2$, \dots $fl10$.
  Furthermore let $\delta(storage1)$ represent the distribution which is non-zero for $storage1$ only. 
  A context that maps the label \densitylabel{LID} to the urn of identifiers and \densitylabel{LStor1} to $\delta(storage1)$ is denoted as follows:
  \begin{align*}
    \context{
      &\density{LID}{\mathcal{U}(fl1,fl2,\dots,fl10)},
      \density{LStor1}{\delta(storage1)}
    }
  \end{align*}
\end{cexpl}

A pair of state formula and valid context forms the basis for a compact abstract state representation (\eref{observation_equivalence} and \eref{state_space_change}), which we call \emph{lifted state}.
A context is valid with respect to a state formula, if and only if 
\begin{inparaenum}[(1)]
  \item each distribution that is referred to within the state formula is defined in the context and
  \item the distribution allows to sample at least as many values as the sum of the cardinalities of the entities.
\end{inparaenum}
A state formula represents a set of grounded states.

\begin{cexpl}\label{expl:lifted_state}
  The state formula from \Explref{state_formula} can be connected to the following context to represent all situations (\ie all permutations of IDs) in that one forklift is at storage room 2 and the other forklifts are in storage room 1:
  \begin{align*}
    \stateform{
      &\entity[9]{\slot{loc}{LStor1}, \slot{ID}{LID}},
      \entity[1]{\slot{loc}{LStor2}, \slot{ID}{LID}}
    }\\
    \context{
      &\density{LID}{\mathcal{U}(fl1,fl2,\dots,fl10)},
      \density{LStor1}{\delta(storage1)},\\
      &\density{LStor2}{\delta(storage2)}
    }
  \end{align*}
\end{cexpl}

Obviously, distributions are not restricted to discrete domains.
The same abstraction can be performed in continuous domains, where lifted states represent infinitely many grounded states.

\subsection{Transition Model}\label{sec:transition_model}

Adapting the idea of computational action languages from CSSMs, our algorithm uses \emph{action schemas} to describe transitions between states (\eref{prediction} and \eref{state_space_change}).
Furthermore, as multiple actions can be executed in parallel (\eref{transitions}), the transitions in the state space actually involve possibly multiple action schemas that are combined to compound actions, similar to those in MRS.

An action schema in LiMa is a triple of
\begin{inparaenum}[(1)]
  \item the \emph{action name} which is necessary for plan recognition,
  \item a \emph{tuple of preconditions} that is used to filter entities that satisfy these conditions, and
  \item an \emph{effect function} mapping entities to a resulting set of entities.
\end{inparaenum}

Preconditions are modeled by constraints over slots that must hold in order to apply the action.
If an action is applied to a set of entities that satisfy the preconditions, the entities within this set are consumed and replaced by the effect of the action.

\begin{cexpl}\label{expl:action_schema}
  The following action schema captures the lifting of a heavy weight box which can only be conducted by \emph{fl1}.
  The precondition makes sure that the value of the slot \slotn{ID} is $fl1$.
  In the effect, a new slot \slotn{load} is created and the value set to $heavy$:
  \begin{align*}
    (&\text{`fl1\_lift\_heavy'},
		&&\text{/* name */}\\
    &(\{\slotn{ID} \mapsto (\lambda v \mapsto v \equiv fl1)\}),
		&&\text{/* precondition */}\\
    &(\ent) \mapsto \multiset{1\ent\{\slotn{load}\mapsto heavy\}})
		&&\text{/* effect */}
  \end{align*}
\end{cexpl}

An action is applicable in a given lifted state if and only if the state contains entities which satisfy the actions preconditions, (\ie, all constraints need to be satisfied by the entity, respectively, the lifted state).
However, if a lifted state contains an entity with uncertain property values, \eg an urn, the preconditions are indeterminate.
Similar to LI, the lifted state is split to decide on the preconditions.
A split is the partition of a lifted state into two lifted states:
\begin{inparaenum}[(1)]
  \item a lifted state that encodes all situations satisfying the preconditions (action can be executed), and
  \item a lifted state encoding all remaining situations (action cannot be executed).
\end{inparaenum}
As part of a split, both the state formula and the context of the lifted state have to be adjusted.
Thus, splits increase the number of hypotheses.
However, to keep the number of hypotheses small, indistinguishable states can be merged.
Merging is the inverse operation of a split.
Note that both concepts are known from the domain of LI.

\begin{cexpl}\label{expl:split}
  Considering the lifted state in \Explref{lifted_state} and the action schema in \Explref{action_schema}, we need \emph{fl1} to satisfy the precondition. 
  The state has to be split into two lifted states:
  \begin{inparaenum}[(1)]
    \item \emph{fl1} being at storage room 1 and
    \item \emph{fl1} being at storage room 2.
  \end{inparaenum}
  In order to perform this split, the urn \densitylabel{LID} is divided into two distributions.
  The first being \densitylabel{LID'}, which represents \densitylabel{LID} without $fl1$ and the second \densitylabel{LID''} which represents $fl1$.
  The resulting states are as follows:
      \begin{align*}
        (1)\quad \stateform{
          &\entity[8]{\slot{loc}{LStor1}, \slot{ID}{LID'}},
          \entity[1]{\slot{loc}{LStor1}, \slot{ID}{LID''}},\\
          &\entity[1]{\slot{loc}{LStor2}, \slot{ID}{LID'}}
        }\\
        \context{
          &\density{LID'}{\mathcal{U}(fl2,\dots,fl10)},\density{LID''}{\delta(fl1)},\\
          &\density{LStor1}{\delta(storage1)},\density{LStor2}{\delta(storage2)}
        }\\
        (2)\quad \stateform{
          &\entity[9]{\slot{loc}{LStor1}, \slot{ID}{LID'}},
          \entity[1]{\slot{loc}{LStor2}, \slot{ID}{LID''}}
        }\\
        \context{
          &\density{LID'}{\mathcal{U}(fl2,\dots,fl10)},\density{LID''}{\delta(fl1)},\\
          &\density{LStor1}{\delta(storage1)},\density{LStor2}{\delta(storage2)}
        }
      \end{align*}
\end{cexpl}

Action application is similar as in MRSs.
Action schemas can be applied to different subsets of entities within the same lifted state.
A set of action schemas can be applied in parallel.
Similarly, compound actions are used to describe sets of actions that are executed in parallel.
Here, we use the concept of maximal parallel compound actions~\citep{barbuti_roberto_maximally_2011}, \ie no further action can be applied in a lifted state.

\subsection{Inference}\label{sec:inference}

A lifted state models multiple situations.
However, as there are several sources of uncertainty (observations, non-deterministic actions, \dots), we will consider not just a single lifted state, but a probability distribution over lifted states, which we call \emph{lifted belief state}.

\begin{cexpl}\label{expl:lifted_belief_state}
  As an example, the following lifted belief state describes two hypotheses.
  One reflects \Explref{lifted_state} with the probability of 0.75 and the other encodes that all forklifts are in storage room 1 with probability 0.25.
  \begin{align*}
    (3)\quad & 0.75 \times
    \stateform{
      \entity[9]{\slot{loc}{LStor1}, \slot{ID}{LID}},
      \entity[1]{\slot{loc}{LStor2}, \slot{ID}{LID}}
    }\\
    &\context{
      \density{LID}{\mathcal{U}(fl1,fl2,\dots,fl10)},
      \density{LStor1}{\delta(storage1)},\\
      &\density{LStor2}{\delta(storage2)}
    }\\
    (4)\quad & 0.25 \times 
    \stateform{
      \entity[10]{\slot{loc}{LStor1}, \slot{ID}{LID}}
    }\\
    &\context{
      \density{LID}{\mathcal{U}(fl1,fl2,\dots,fl10)},
      \density{LStor1}{\delta(storage1)}
    }
  \end{align*}
\end{cexpl}

Starting with an initial lifted belief state, in the framework of BF (\eref{sequential}), we need to perform consecutively
\begin{inparaenum}[(1)]
  \item updating the belief state according to the observations (\eref{update}),
  \item answering of application specific questions, and
  \item predicting the next lifted belief state by applying the compound actions.
\end{inparaenum}
Note that all updates and manipulations of the belief states are performed in the lifted domain without grounding.

While applying the observation update, splits are required if identifying observations are made.
Application specific questions can be answered on every updated posterior distribution over lifted states.

\begin{cexpl}
  We are interested in the location of \emph{fl1} for the lifted belief state in \Explref{lifted_belief_state}.
  To determine the location of \emph{fl1} we need to split both lifted states:
  \begin{compactenum}
    \item 
      Splitting state (3) leads to two lifted states (1) and (2) from \Explref{split}. 
      Whereas the first has 9 different instantiations, the second has a single one, which naturally lead to the probabilities of $0.9$ and $0.1$, respectively. 
      Weighted by the probability of the original lifted state (3) this results in $0.675$ and $0.075$.
    \item 
      From (4) we know that \emph{fl1} is at location storage room 1 with probability $0.25$. 
  \end{compactenum}
  Summing the probabilities, leads to a probability of $0.675+0.25=0.925$ that \emph{fl1} is at storage room 1 and $0.075$ that it is at storage room 2.
\end{cexpl}

\section{Experiments}\label{sec:Experiments}

\begin{figure}[tb]
  \centering
  \includegraphics[width=0.48\textwidth]{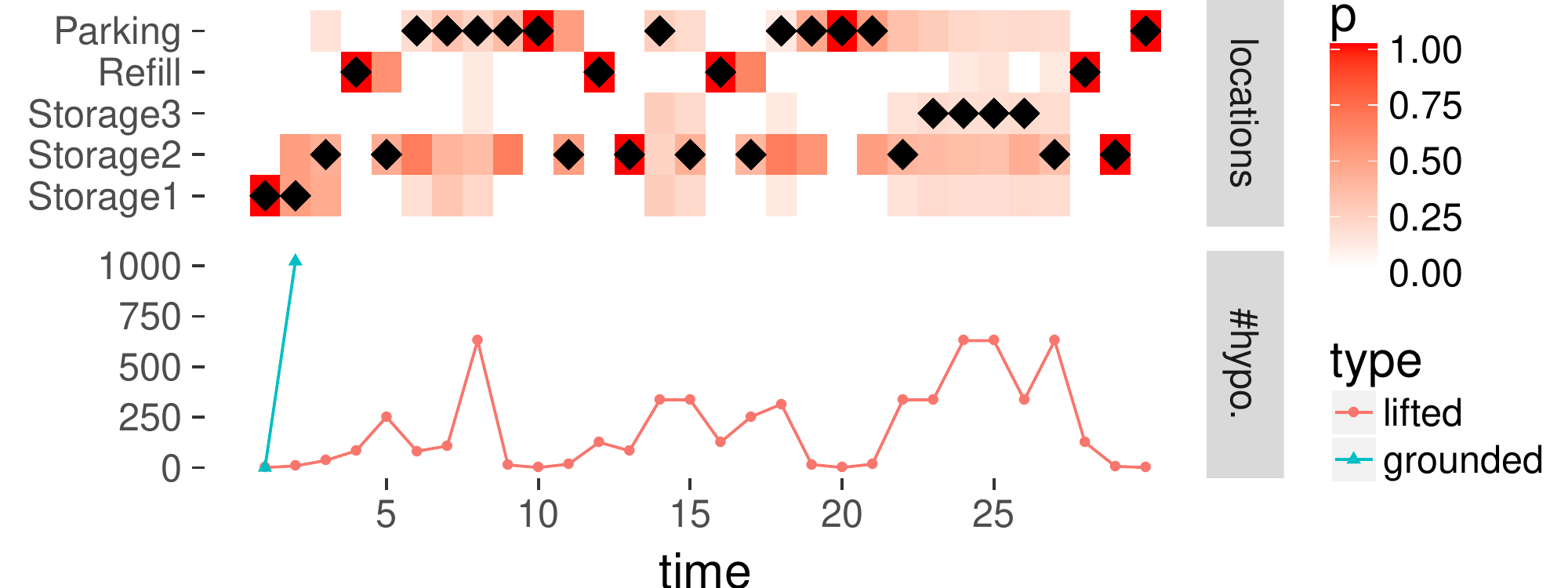}
  \caption{
  	Results for the warehouse scenario. 
  	Top: Estimation of the location of \emph{fl1}, black diamonds: true location. 
		Bottom: Number of hypotheses during inference. Notice that the number of hypotheses gets small when all forklifts cluster at few locations (time 10, 20, 30). Identifying observations are made at the refill location. Grounded inference was infeasible after timestep 2.
  }
  \label{fig:nhresults}
\end{figure}

\begin{figure}[tb]
  \centering
  \includegraphics[width=0.48\textwidth]{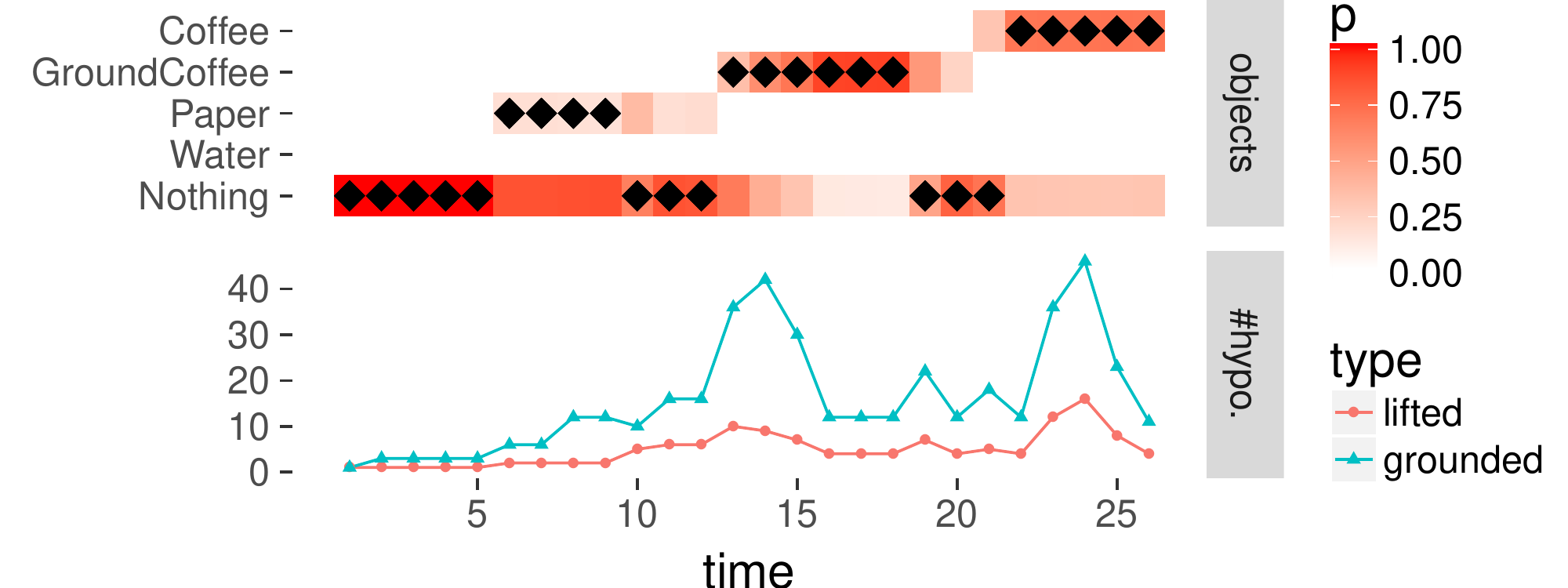}
  \caption{
    Results for the office scenario. 
    Top: Estimation of the object person 1 is holding, black diamonds: true object.
    Bottom: Number of hypotheses during inference. An identifying observation has been made at timestep 10.}
  \label{fig:abcresults}
\end{figure}

In this section, we evaluate the Lifted Marginal Filtering algorithm on the two scenarios (\secref{Application_Requirements}). 
\Scnrref{warehouse} is a location tracking task, \Scnrref{office} has a more structured state space with additional entity information. 
Both scenarios are used to compare our approach with a grounded inference approach.
For our experiments we randomly sampled observation sequences from a HMM.
Performance is assessed by counting the number of hypotheses during inference.
Furthermore, application specific questions are answered.

\subsection{Warehouse Scenario}

As can be seen from upper part of \figref{nhresults}, the location of forklift \emph{fl1} can be determined exactly when either an identifying observation was made (time 4, 12, 16, and 28), or all forklifts are at the same location (time 10, 20, and 30).
Knowledge about such situations is incorporated in preceding timesteps, as for instance one timestep after identification, \emph{fl1} cannot be at storage room 1, because it takes two time steps to get there (cf. time 4 and 5 in \figref{nhresults}).

\figref{nhresults} (lower part) shows the number of hypotheses tracked during inference.
At most, 630 hypotheses had to be tracked simultaneously in our algorithm.
The corresponding grounded inference algorithm was infeasible after timestep 2 because of the large number of hypotheses.
Specifically, the algorithm exceeded the available 128 GB memory.
The number of hypotheses decreases one timestep before all forklifts meet at the same location (time 10, 20, 30), and one timestep after this event.
The reason is that there are only two possible locations for the forklifts (parking room and storage room 2).
This situation can be represented by very few hypotheses in our algorithm.

This experiment shows that LiMa can efficiently perform BF in situations with mixed anonymous and identifying observations.
Grounded inference, in contrast, is infeasible.
It is worth mentioning that LiMa performs \emph{exact} inference, as opposed to IM approaches that use approximation.

\subsection{Office Scenario}

The office scenario requires to model more complex situations that go beyond location tracking tasks, since different entity properties (what a person holds) and different types of entities (persons and printjobs) are involved.
\Figref{abcresults} depicts the the number of hypotheses during inference and as application specific questions, the object that person 1 holds.
The maximum number of hypotheses of our approach is 16, compared to 46 hypotheses tracked by the grounded approach. 

We restricted the number of persons to 3 to enable the comparison with a grounded version. 
Since LiMa uses an abstract state representation, it would have been able to handle larger numbers.

This scenario shows that LiMa is not only capable of tracking object locations, but can also handle scenarios that contain much more internal structure, e.g. multiple entity properties and multiple different actions.

\section{Conclusion}\label{sec:Conclusion}
We introduced a general version of the \emph{data association problem} and showed that it occurs in many real world scenarios.
In this problem class, multiple similar entities act (or get manipulated) simultaneously, while the observations allow only incomplete knowledge about their identity.
This leads to a combinatorial explosion of the number of hypotheses.
We showed that none of the state of the art approaches satisfies all of the requirements resulting from the generalized version of this problem.
However, ideas from different state of the art methods can be combined to a novel filtering algorithm that can handle this problem, which we call \emph{Lifted Marginal Filtering}.
The state space representation, and the description of state transitions is based on Multiset Rewriting.
Furthermore, the entities are structured, such that additional properties of an entity, like its location or name, can be modeled. 
Based on the presence of certain information of entity properties, entities are either represented explicitly or grouped together by representing their properties parametrically.
We can dynamically modify the granularity of these representations by \emph{splitting} and \emph{merging} abstract states.
Two scenarios were used to illustrate the superior state space representation.

Future work includes the automatic switching from exact to approximate inference if the state space gets to large.
Another aspect is the investigation of approximate merging, that is, grouping multiple states even if the resulting state does not exactly represent the same distribution.
A further research goal is to investigate whether entity multiplicities can be represented parameterically. 

\bibliographystyle{Template/aaai}
\bibliography{P20170622_STARAI_Requirements_MS}

\end{document}